\documentclass[10pt,journal,compsoc]{IEEEtran}

\ifCLASSOPTIONcompsoc
  \usepackage[nocompress]{cite}
\else
  \usepackage{cite}
\fi

\ifCLASSINFOpdf
\else
\fi
\usepackage{helvet}
\usepackage{courier}
\usepackage{amsmath}
\usepackage{amssymb}
\usepackage{graphicx}
\usepackage{subfigure}
\usepackage{longtable}
\usepackage{rotating}
\usepackage{multirow}
\usepackage{multicol}
\usepackage{xcolor}
\usepackage{fancyhdr}
\usepackage{enumerate}
\usepackage{float}
\usepackage{url}
\usepackage[linesnumbered,ruled]{algorithm2e}
\usepackage{wrapfig}

\usepackage{balance}
\usepackage{makecell}
  \newcommand\figcaption{\def\@captype{figure}\caption}
  \newcommand\tabcaption{\def\@captype{table}\caption}
\makeatother
\begin{document}

\title{Generative One-Shot Face Recognition}

\author{Zhengming~Ding,
Yandong~Guo,
Lei~Zhang,~\IEEEmembership{Senior Member, IEEE},
        and~Yun~Fu,~\IEEEmembership{Senior Member, IEEE}
\IEEEcompsocitemizethanks{\IEEEcompsocthanksitem Z. M. Ding is with the Department of Computer, Information and Technology
Indiana University-Purdue University Indianapolis, 420 University Blvd Indianapolis, IN 46202.\protect\\
E-mail: allanzmding@gmail.com

\IEEEcompsocthanksitem Yandong Guo is with XPeng Motors,\protect\\
      Email: yandong.guo@live.com

\IEEEcompsocthanksitem Lei Zhang is with Microsoft Research,\protect\\
      Email: leizhang@microsoft.com

\IEEEcompsocthanksitem Y. Fu is with the Department of Electrical and Computer Engineering and the College of Computer and Information Science, Northeastern University, Boston, MA, 02115 USA.\protect\\
E-mail: yunfu@ece.neu.edu.}}

\markboth{IEEE TRANSACTIONS ON PATTERN ANALYSIS AND MACHINE INTELLIGENCE}%
{Shell \MakeLowercase{\textit{et al.}}: Bare Demo of IEEEtran.cls for Computer Society Journals}

\IEEEtitleabstractindextext{%
\begin{abstract}
One-shot face recognition measures the ability to identify persons with only seeing them at one glance, and is a hallmark of human visual intelligence. It is challenging for conventional machine learning approaches to mimic this way, since limited data are hard to effectively represent the data variance. The goal of one-shot face recognition is to learn a large-scale face recognizer, which is capable to fight off the data imbalance challenge. In this paper, we propose a novel generative adversarial one-shot face recognizer, attempting to synthesize meaningful data for one-shot classes by adapting the data variances from other normal classes. Specifically, we target at building a more effective general face classifier for both normal persons and one-shot persons. Technically, we design a new loss function by formulating knowledge transfer generator and a general classifier into a unified framework. Such a two-player minimax optimization can guide the generation of more effective data, which effectively promote the underrepresented classes in the learned model and lead to a remarkable improvement in face recognition performance. We evaluate our proposed model on the MS-Celeb-1M one-shot learning benchmark task, where we could recognize $94.98\%$ of the test images at the precision of $99\%$ for the one-shot classes, keeping an overall Top1 accuracy at $99.80\%$ for the normal classes. To the best of our knowledge, this is the best performance among all the published methods using this benchmark task with the same setup, including all the participants in the recent MS-Celeb-1M challenge at ICCV 2017\footnote{http://www.msceleb.org/challenge2/2017}.
\end{abstract}

\begin{IEEEkeywords}
One-Shot Learning, Generative Adversarial Nets, Large-Scale Face Recognition
\end{IEEEkeywords}}

\maketitle

\IEEEdisplaynontitleabstractindextext

\IEEEpeerreviewmaketitle

\IEEEraisesectionheading{\section{Introduction}\label{sec:introduction}}

\begin{figure}[t]
\begin{center}
      \includegraphics[width=0.49\textwidth]{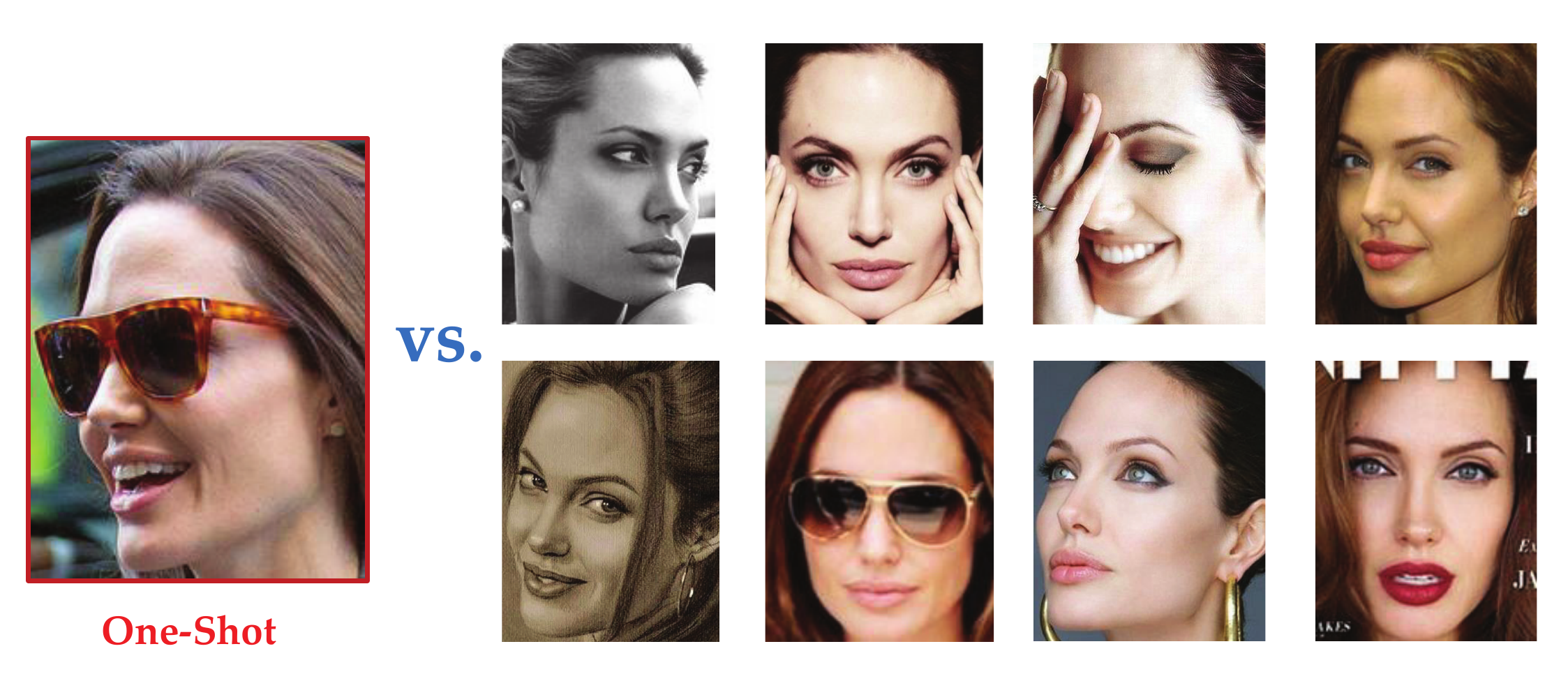}
  \vspace{-7mm}\caption{Illustration of One-Shot Challenge, where the one-shot image in the leftmost column is used for training and the rest images (in the right panel) are the corresponding images for testing (partially selected from the test set). With only one image for each person, the challenge is how to recognize all these test images from hundreds of thousands of other testing images. More detailed results are presented in the experimental results section. 
}\vspace{-6mm}
\label{problem}
\end{center}
\end{figure}

\IEEEPARstart{O}{ne-shot} face recognition is to recognize persons with only seeing them once (Figure \ref{problem}). This problem exists in many real applications. For example, in the scenario of large-scale celebrity recognition, it naturally happens that some celebrities only have one or very limited number of images available. Another example is in the law enforcement scenario: it is usually the case that only one image of the personal ID is available for the target person. The challenge of one-shot face recognition lies in two parts. 

\emph{First of all}, a representation model is needed to transfer the face image into a discriminative feature domain. Although recent years have witnessed great progresses in deep learning for visual recognition, computer vision systems still lack the capability of learning visual concepts from just one or a very few examples \cite{lake2015human}. A typical solution is to leverage many images from a different group of people (we call them \textit{base set} and name the persons with limited number of training images \textit{low-shot set}), and train a representation model using the images from the base set to extract face features for the images in the low-shot set. Recently, there have been many research efforts focusing on training representation models with good generalization capability \cite{FaceBook_2014,FaceBook_2015,schroff2015facenet,vgg_face}, where face representation model is trained and tested across different groups of persons. However, improving the generalization and capability of face representation model is still an open problem which has attracted substantial effort in the area. When the distributions of the face dataset-A and face dataset-B are very different, the representation model trained on dataset-A may not be discriminative enough on dataset-B. For example, if the data used to train a representation model do not include sufficient number of images for persons with a certain type of skin color, the trained model usually suffer from lower accuracy for those persons.

\begin{figure*}[t]
\begin{center}
      \includegraphics[width=0.9\textwidth]{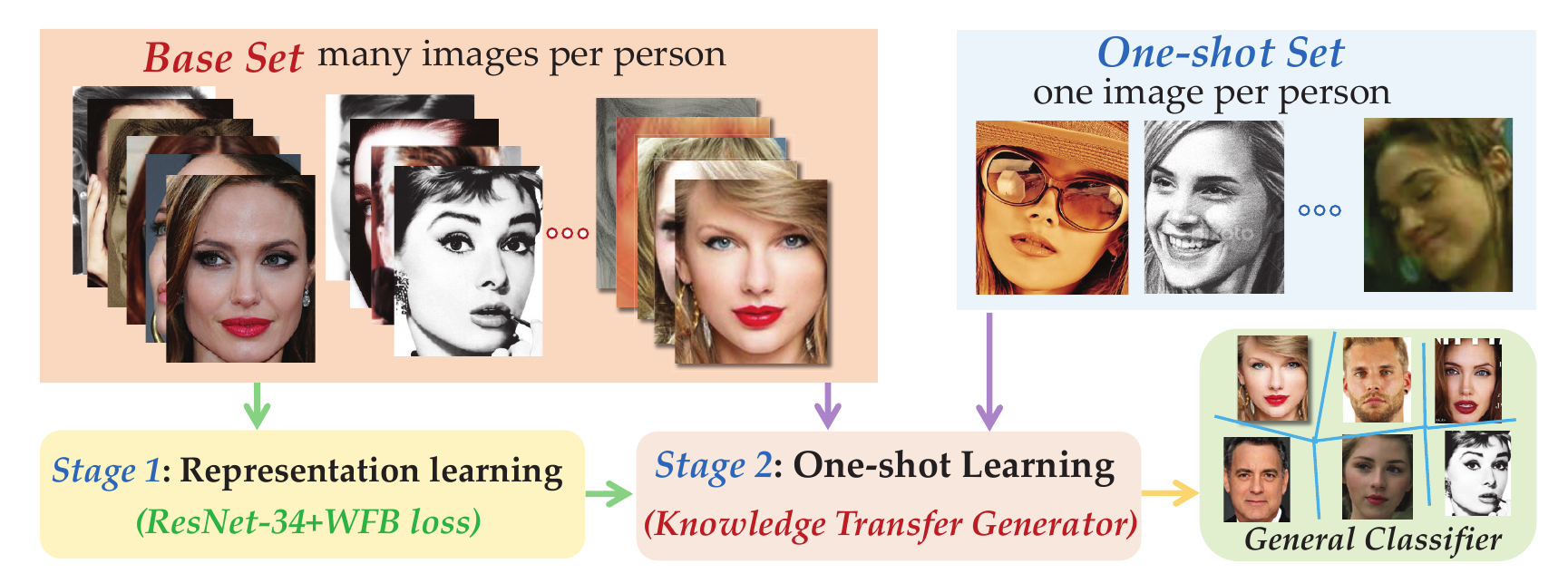}
  \vspace{-3mm}\caption{Illustration of one-shot face recognition problem with two phases. Stage 1: \textbf{Representation learning} phase seeks general face visual knowledge through training effective feature extractor using the base set. Stage 2: \textbf{One-shot GAN learning} phase builds a general classifier to recognize persons in both the base set and the one-shot set based on the deep features. }\vspace{-5mm}\label{framework}
\end{center}
\end{figure*}

\emph{Secondly}, the challenge of one-shot face recognition comes from estimating the partition for a given person in the feature space. A representation model transfers the face images of the same person into a cluster of dots in the feature space. To recognize all the faces for a given person, we need to estimate the shape, size, and location of the partition for this person in the feature space. However, with only one image (corresponding to one dot in the feature space), it is not easy to accurately and reliably estimate the distribution of the faces of the person to be recognized, which makes it challenging to estimate the boundary of the partition for this person in the feature space. A straightforward yet a bit over-simplified way to estimate  the partition for the one-shot person in the feature space is to assume each person claims a hyper-sphere with equal size in the feature space, and then use the $k$-nearest neighborhood ($k$-NN) classifier with a certain threshold to recognize persons each having only one image available. However, many recent works \cite{guo2017one,Wu_2017_ICCV,wu2016deep} demonstrate that directly using $k$-NN is often a sub-optimal solution compared with other methods which can learn the partition boundary in more informative way. 

In order to better study the one-shot face recognition problem, there have been some benchmark tasks designed. One typical example is MegaFace in \cite{UW_MegaFace}. In this benchmarks task, one face image for a given celebrity is provided to search for another face image for this celebrity 
from up-to one million distractor images from regular persons. Since this task requires participants to extract feature vectors for all the face images and use 1-NN method to search for the image with the smallest distance, this task mainly focuses on evaluating the first challenge of the one-shot face recognition: to learn a discriminative face representation model. Recently, there has been another benchmark task called MS-Celeb-1M: low-shot challenge \cite{guo2017one} proposed. This task focuses on one-shot learning in the large-scale face recognition scenario. The task is to study, with these training images only, how to develop an algorithm to recognize the persons in \textbf{both} data sets. The main focus of this task is the recognition accuracy for persons in the low-shot set as it shows the one-shot learning capability of a vision system, while also checking the recognition accuracy for those in the base set to ensure no harm to their performance.

Most recently, one-shot learning has attracted great attentions, attempting to make progress towards imparting this human ability to modern recognition systems \cite{hariharan2016low,wang2016learning,mehrotra2017generative,motiian2017few,guo2017one,gidaris2018dynamic,wang2018low,bartunov2018few}. Generally, there are two ways to improve the low-shot face recognition performance. The first strategy is to enhance the generalization and discriminative capability of representation model. Examples include range loss \cite{RangeLoss}, fisher face \cite{Fisher}, center invariant loss \cite{centerinvariant}, marginal loss \cite{MarginalLoss}, sphere face \cite{sphereface}, etc. The second strategy is to improve the estimation of partitions in the feature space, which mainly includes two lines, i.e., data augmentation and classifier adaptation. Along the first line, some human-designed data generation strategies are adopted to synthesize fake data for the low-shot/one-shot classes to boost the classification ability. Edwards et. al proposed handling the one-shot classification task by learning dataset statistics using the amortized inference of a variational auto-encoder \cite{edwards2016towards}. Hariharan et al. designed a method for hallucinating additional examples for the data-starved novel classes \cite{hariharan2016low}. Mehrotra et al. presented an additional generator network based on the Generative Adversarial Networks where the discriminator is the proposed residual pairwise network \cite{mehrotra2017generative}. For the second line, the idea is to adapt the base classifier to the novel classifier through multi-layer transformations or some specific loss to boost the classifier space of one-shot classes. Guo et al. proposed a novel supervision loss named as Underrepresented-classes Promotion (UP) loss term, which aligns the norms of the weight vectors of the one-shot classes to those of the normal classes \cite{guo2017one}, which directly boosts the one-shot classifier parameters  without considering the difference of class variance for different one-shot classes. However, human-designed generation rules cannot learn the data distribution well to synthesize data to effectively improve the recognition. 

In this work, we address one-shot face recognition includes the following two phases (Figure \ref{framework}). The first phase is named as \textit{representation learning}. In this phase, we build face representation model using all the training images from the \textit{base set}. The second phase is called as \textit{one-shot learning}. In this phase, we train a multi-class classifier to recognize the persons in both \textit{base set} and \textit{one-shot set} based on the representation model learned in phase one. We design a generative one-shot learning model to improve the recognition performance for the persons in the one-shot set. The core idea of our generate model is to synthesize effective auxiliary data for the one-shot classes, and thus we can span the feature space for the one-shot classes to facilitate the one-shot face recognition. To our best knowledge, this is one of the first works to explore the generative model in a general classifier learning for one-shot challenge. Specifically, we mainly focus on the one-shot learning stage, where we attempt to train a more powerful classifier for both base and novel classes. Finally, we also provide several interesting and challenging future directions for one-shot face recognition. The main contributions of our paper are listed in three folds as follows:
\begin{itemize}
  \item We jointly incorporate generative adversarial networks in training a general classifier for both base and novel classes. In detail, the generator attempts to synthesize more effective fake data for the one-shot classes to enrich the data space of one-shot classes, while the discriminator is built to guide the face data generation to mimic the data variation of base classes and adapt to generate novel classes. 
  \item We design generative adversarial networks in the feature domain which we first obtain by training a deep ConvNet model on the base classes \footnote{We train generative model on feature domain instead of image domain for the following reasons. Typically, image synthesis is a more challenging task than image classification/recognition. Especially for the current generative models, it is still an open problem to generate meaningful faces with high quality in many cases \cite{mirza2014conditional,chen2016infogan}. This might be the reason that we don't find any existing one-shot learning work using generative models to synthesize face images. Furthermore, we joint our generative model and the deep architecture into a unified model to seek more general feature extractor.b}. More specifically, we build the conditional generative adversarial networks with an auxiliary classifier to augment more effective features and enhance the general classifier learning for one-shot classes. 
  \item We evaluate our proposed model on a large-scale one-shot face dataset, and achieve significant improvement in the one-shot classification with coverage rate $94.98\%$ at the precision of $99\%$. Meanwhile, our model can still achieve very appealing performance as Top1 accuracy of $99.80\%$ for the base classes.
\end{itemize}

The rest sections of this paper are organized as follows. In Section 2, we first review the progress of one-shot face recognition nowadays, later present a brief discussion of the related works and highlight the difference between zero-shot learning, transfer learning and one-shot learning. Then we propose two-stage one-shot learning including representation learning and one-shot face recognizer through generative learning in Section 3. Experimental evaluations on large-scale face datasets are reported in Section 4, which is followed by the conclusion in Section 5.

\section{Related Work}

In this section, we mainly review the current status of one-shot learning in the literature. 

In the general image recognition domain, the recent low-shot learning work \cite{hariharan2016low,wang2018low} also attracts a lot of attentions. Their benchmark task is very similar to one-shot face recognition but in the general image recognition domain: the authors split the ImageNet data\footnote{http://www.image-net.org/} into the base and low-shot (called novel in \cite{hariharan2016low}) classes, and the target is to recognize images from both the base and low-shot classes. Their solution is quite different from ours since the domain is quite different. We will not review their solution here due to the space constraint, but list results from their solution as one of the comparisons in the experiments section. 

Overall, one-shot learning is still an open problem. A natural source of information comes from additional data via ``data manufacturing'' \cite{bart2005cross} in various ways. In a broad sense, learning novel classes is addressed by exploiting and transferring knowledge gained from familiar classes. This is to imitate the human ability of adapting previously acquired experience when recognizing novel classes. In the following, we will revisit different categories of one-shot learning methods, including the most popular ``general feature learning'', classifier learning \& adaption, and data augmentation. 

\subsection{Generalized Feature learning}

Cross entropy with Softmax has demonstrated good performance in supervising the face feature extraction model training. In order to further improve the performance of representation learning, many methods have been proposed to add extra loss terms or slightly modify the cross entropy loss (used together with Softmax for multinomial logistic regression learning) to regularize the representation learning in order to improve the feature discrimination and generalization capability. 

Among all these works, we consider the center loss \cite{wen2016discriminative} as one of the most representative one (a similar idea published in \cite{Latha:dense} during the same time). In \cite{wen2016discriminative}, face features from the same class are encouraged to be close to their corresponding class center (actually, approximation of the class center, usually dynamically updated). By adding this loss term to the standard Softmax, the authors obtain a better face feature representation model \cite{wen2016discriminative}. There are many other alternative methods, including the range loss in \cite{RangeLoss}, fisher face in \cite{Fisher}, marginal loss in \cite{MarginalLoss}, sphere face in \cite{sphereface}, etc. Each of these methods has its own uniqueness and advantages under certain setup. Guo et al. designed a different kind of loss term adding to the cross entropy loss of the Softmax to improve the feature extraction performance \cite{guo2017one}. Compared with center loss in \cite{wen2016discriminative} or sphere face in \cite{sphereface} (these two are the most similar ones), Guo et al. demonstrated that their proposed method has better performance from the perspective of theoretical discussion and experimental verification. Unfortunately, it is not very practical to compare all these cost function design methods fairly and thoroughly, since these cost functions were implemented with different networks structures, and trained on different datasets. Sometimes parameter adjustment is critically required when the training data is switched. 

\subsection{$k$-Nearest Neighbor \& Softmax Classifier}

After a good face feature extractor is obtained, the template-based method, e.g., $k$-nearest neighborhood ($k$-NN) classifier, is widely used for face identification these days. The advantages of $k$-NN is clear: no classifier training is needed, and $k$-NN does not suffer much from imbalanced data, etc. However, experiments in \cite{wu2016deep,Xu_2017_ICCV,guo2017one,Wu_2017_ICCV} demonstrate that the accuracy of $k$-NN with the large-scale face identification setup is usually lower than Softmax Classifier, when the same feature extractor is used. Moreover, if we use all the face images for every person in the gallery, the complexity is usually too high for large scale recognition, and the gallery dataset needs to be very clean to ensure high precision. If we do not keep all the images per person, how to construct representer for each class is still an open problem. 

As described above, Softmax classifier demonstrates overall higher accuracy compared with $k$-NN in many previous publications. This is mainly because in Softmax classifier, the weight vectors for each of the classes is estimated using discriminant information from all the classes, while in the $k$-NN setup, the query image only needs to be close enough to one local class to be recognized. Moreover, after feature extraction, with Softmax classifier, the computational complexity of estimating the persons' identity is linear to the number of persons, not the number of images in the gallery. However, the standard Softmax classifier suffers from the imbalanced training data and has poor performance with the low-shot classes even these classes are oversampled, though the overall accuracy is higher than $k$-NN. Recently, some works develop hybrid solutions by combining Softmax classifier and $k$-NN \cite{Xu_2017_ICCV,Wu_2017_ICCV} and achieve promising results. In these work, when Softmax classifier does not have high confidence (threshold tuning is needed), $k$-NN is used. 

We solve this problem from a different perspective. Different from the hybrid solution, our solution only has one Softmax classifier as the classifier so that no threshold is needed to switch between classifiers. We boost the performance of Softmax classifier by involving the generated data.

\subsection{Classifier Adaptation}

Another type of knowledge transfer focuses on modeling (hyper-)parameters that are shared across domains, typically in the context of generative statistical modeling \cite{fe2003bayesian,lee2007learning,rodner2010one}. Li et al. operated in a variational Bayesian framework by incorporating previously learned classes into the prior and combining with the likelihood to yield a new class posterior distribution \cite{fe2003bayesian,fei2006one}. Gaussian processes and hierarchical Bayesian models are also employed to allow transferring in a non-parametric Bayesian way. Specifically, hierarchical Bayesian program learning utilizes the principles of compositionality and causality to build a probabilistic generative model of visual objects \cite{lake2013one,lake2015human}. In addition, adaptive SVM and its variants present SVM-based model adaptation by combining classifiers learned on related categories \cite{wang2016learning,tommasi2014learning}. Wang et al. assumed there exists a generic, category agnostic transformation from small-sample models to the underlying large-sample models \cite{Wang2016a}, and thus they explored a novel learning to learn approach that leverages the knowledge gained when learning models in large sample sets to facilitate recognizing novel categories from few samples. Despite many notable successes, it is still unclear what kind of underlying structures are shared across a wide variety of categories and are useful for transfer.

\subsection{Data Augmentation}

Data augmentation is a straightforward way to boost the one-shot class performance, since it could compensate the shortage of one-shot class samples by synthesizing more data. However, it is the key to generate meaningful data with enough data variance for one-shot classes. Along this line, Hariharan et al. presented a way of ``hallucinating'' additional examples for one-shot (low-shot) classes by transferring modes of variation from the base classes \cite{hariharan2016low}. Their experiments demonstrated those additional examples improve the one-shot top-5 accuracy on low-shot classes while also maintaining accuracy on the base classes. Note that Hariharan et al. adopted some human-designed rules to augment the data space for low-shot classes, which is very complexed and not easily spread in real-world applications \cite{hariharan2016low}. 

Most recently, generative models \cite{goodfellow2014generative,mirza2014conditional,dai2017good} are exploited to synthesize more training data for one-shot classes by automatically capturing the data variance from base classes. Specifically, Rezende et al. developed a class of sequential generative models by combining the representational power of deep learning with the inferential power of Bayesian reasoning, which is among the state-of-the art in density estimation and image generation \cite{rezende2016one}. Choe et al. adapted a generator to increase the size of training dataset, which includes a base set, a widely available dataset, and a novel set, a given limited dataset, while adopting transfer learning as a backend \cite{ICCV-W-Generation}. Mehrotra et al. proposed a deep residual network with an additional generator network that allows the efficient computation of this more expressive pairwise similarity objective \cite{mehrotra2017generative}. Our proposed generative model also belongs to this category. The key difference from previous work is we exploit data generation in feature domain instead of image domain by jointly seeking a general classifier. Moreover, we involve the class center and class variance to synthesize more efficient data. Moreover, this work is our previous conference extension \cite{ding2018oneshot}. In this journal extension, we review more recent progress on one-shot learning. Then, we improve our representation learning model with WFB loss (Section 3.2), in term of the feature discrimination and generalization capability. The newly proposed WFB loss encourages the face features belong to the same class to have similar direction as their associated classification weight vectors. Moreover, we further enhance our generative model by incorporating the data variance of base classes. Our generator can better mimic the human cognitive by transferring their previous knowledge. Here the previous knowledge denotes the data variance from base classes.

\section{The Proposed Algorithm}

The objective of one-shot face recognition is to measure the recognition ability of a model across classes with one training sample. Specifically, the model is trained on labeled training data with two sets without identity overlap, i.e., \textit{base set} (i.e., normal classes) $\{X_b, Y_b\}$ with $c_b$ classes and \textit{one-shot set} $\{X_n, Y_n\}$ with $c_n$ classes. The goal is to build a general $c$-class recognizer $(c= c_b+c_n)$. We address one-shot face recognition includes the following two phases (Figure \ref{framework}). The first phase is named as \textit{representation learning}. In this phase, we build face representation model using all the training images from the \textit{base set}. The second phase is called as \textit{one-shot learning}. In this phase, we train a multi-class classifier to recognize the persons in both \textit{base set} and \textit{one-shot set} based on the representation model learned in phase one. We design a generative one-shot learning model to improve the recognition performance for the persons in the one-shot set.  

\subsection{Motivation}

\begin{figure}[t]
\begin{center}
      \includegraphics[width=0.49\textwidth]{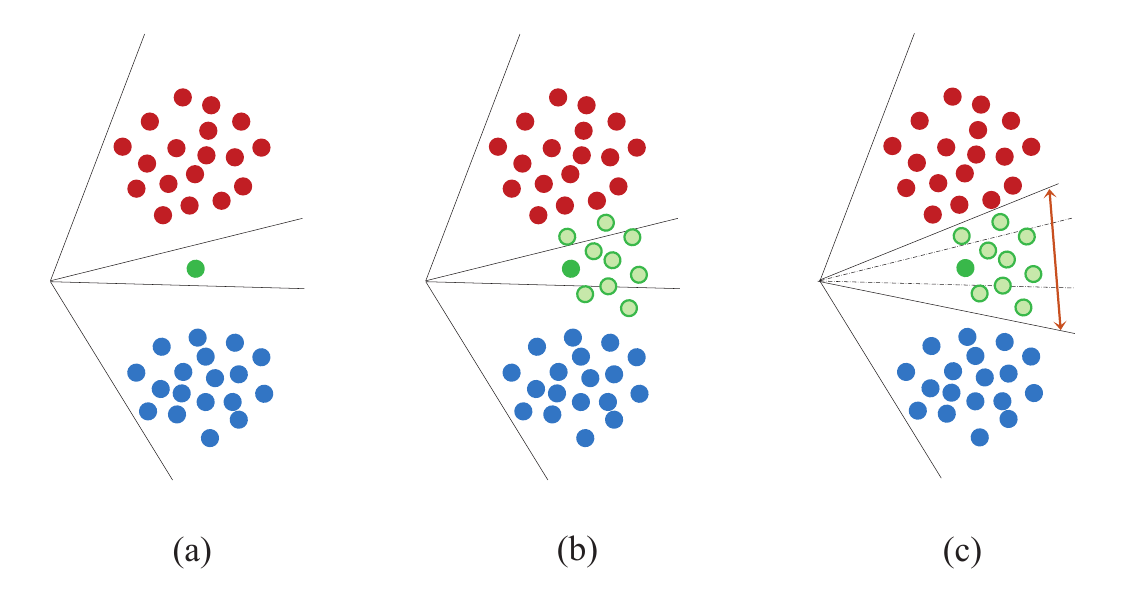}
  \vspace{-8mm}\caption{Illustration of generative model to synthesize more samples for one-shot classes then update decision boundary bias. (a) in the beginning, we have one training sample for one-shot class while many samples for base classes, thus the classifier would be dominated by the bias. (b) we explore our generate model to synthesize more samples for one-shot class. (c) with the augmentation of feature space for one-shot classes, the classifier also be updated with its one-shot classifier space enlarged. }\vspace{-6mm}\label{gc}
\end{center}
\end{figure}

One-shot face recognition is challenging due to limited samples during model training, while general deep frameworks treat base and one-shot class equally, which leads to biased updates of the recognition model. Thus, it is essential to generate more effective data to improve the ability of the general classifier. Traditional data augmentation strategies \cite{hariharan2016low} only adopt human designed rules to generate more data for the one-shot classes. Hence, the enhancement to the classifier is limited. Another challenge is that it usually hurts the base classification when we try to improve the classification ability for one-shot classes. That is, the learned classifier is impractical in real-world applications when dealing with a general face recognition problem. Hence, it is essential to balance these two sets.

Moreover, generative models are very popular due to its promising ability to synthesize effective data automatically, which are similar to the real data with a guidance from a discriminator. While for one-shot face recognition, it is essential to generate effective data with large variations for the one-shot classes in order to span their classifier space. Generally, data from the base classes have large within-class variations, and thus, it is helpful to adapt the variations of base classes to the one-shot classes during data generation. To generate more meaningful data for one-shot classes, we jointly seek a general classifier with the input of real data and fake data. Such a joint learning framework could benefit generating meaningful data and improving the classification ability, specially updating the decision boundary of the classifier shown in Figure \ref{gc}, where more meaningful data are augmented to enlarge the feature space then updating the classifier boundary. 

\subsection{Representation Learning}

To learn more effective feature representation, we design our face representation model with supervised learning framework considering persons' identities as class labels. Specifically, we propose the loss function as follows:
\begin{equation}
\label{eq:costfunction}
\mathcal{L} =  \mathcal{L}_s + \lambda \mathcal{L}_a
 \, ,  
\end{equation}
where $\mathcal{L}_s$ is the standard cross-entropy loss used for the Softmax layer, while $\mathcal{L}_a$ is the newly proposed loss used to improve the feature discrimination and generalization capability. $\lambda$ is the trade-off parameter between two loss functions. More specifically, we recap the first term, cross-entropy $\mathcal{L}_s$ as follows:
\begin{align}
\label{eq:crossentropy}
\mathcal{L}_s &= - \sum\limits_i \sum\limits_k t_{k,i} \log p_{k}(\emph{x}_i),
\end{align}
where $t_{k,i} \in \{0,1\}$ is the ground truth label indicating whether the $i$-{th} image belongs to the $k$-{th} class, and the term $p_{k}(\emph{x}_i)$ is the estimated probability that the image $\emph{x}_i$ belongs to the $k$-{th} class, defined as, 
\begin{equation}
\label{eq:sigmoid}
p_{k}(\emph{x}_n) = \dfrac{\exp \big(\mathbf{{w}}_k^\top {\boldsymbol{\phi}(\emph{x}_i)} \big)}{\sum\limits_i \exp \big(\mathbf{w}_k^\top {\boldsymbol{\phi}(\emph{x}_i)}\big)} \, ,
\end{equation}
where $\mathbf{w}_k$ is the weight vector for the $k$-{th} class, and $\boldsymbol{\phi}(\cdot)$ denotes the feature extractor for  image $\emph{x}_n$. Note that in all of our experiments, we always set the bias term $b_k=0$. We choose the standard residual network with 34 layers (ResNet-34) \cite{Resnet} as our feature extractor $\boldsymbol{\phi}(\cdot)$ using the last pooling layer as the face representation. ResNet-34 is used due to its good trade-off between prediction accuracy and model complexity, yet our method is general enough to be extended to deeper network structures for even better performance. We have conducted comprehensive experiments and found that removing the bias term from the standard Softmax layer in deep convolutional neural network does not affect the performance. However, it is worth noting that this leads to a much better understanding of the geometry property of the classification space.

The second term $\mathcal{L}_a$ in the cost function (Eq. \eqref{eq:costfunction}) is defined as
\begin{align}
\label{eq:cca}
\mathbf{w}'_k &\leftarrow \mathbf{w}_k \\
\mathcal{L}_a &= - \sum_k\sum_{i \in C_k} \frac{\mathbf{w}_k^{'\top} \boldsymbol{\phi}(\emph{x}_i)}{\|\mathbf{w}'\|_2 \|\boldsymbol{\phi}(\emph{x}_i)\|_2} \,,
\end{align}
where we set the parameter vector $\mathbf{w}'_k$ to be equal to the weight vector $\mathbf{w}_k$. This loss term encourages the face features belong to the same class to have similar direction as their associated classification weight vector $\mathbf{w}_k^\top$. We name this loss term as Weights-guided Feature vector Bundling (WFB). Calculating the derivative with respect to $\boldsymbol{\phi}(\emph{x}_i)$, we have 
\begin{equation}
\dfrac{\partial \mathcal{L}_a}{\partial \boldsymbol{\phi}(\emph{x}_i)}=\frac{1}{\|\boldsymbol{\phi}(\emph{x}_i) \|_2} 
\left(
\frac{\mathbf{w}_k^{'\top}}{\|\mathbf{w}'_k\|_2}  
-\frac{\boldsymbol{\phi}(\emph{x}_i)^\top \cos \theta_{i,k}}{\|\boldsymbol{\phi}(\emph{x}_i)\|_2} 
\right) \,,
\end{equation}
where $\theta_{i,k}$ is the angle between $\mathbf{w}'_k$ and $\boldsymbol{\phi}(x_i)$. Note that $\mathbf{w}'_k$ in this term is the parameter copied  from $\mathbf{w}_k$, so there is no derivative to $\mathbf{w}'_k$. For experiment ablation purpose, we also tried to back propagate the derivative of $\mathbf{w}_k$, but did not observe better results. 

To further understand Eq. \eqref{eq:costfunction}, without loss of generality, we discuss the decision hyperplane between any two adjacent classes. Note we set all the bias terms $b_k$ and $b_j$ to $0$ throughout the paper. With this setup, we apply Eq. \eqref{eq:sigmoid} to both the $k$-{th} class and the $j$-{th} class to determine the decision hyperplane between the two classes (note we do not have bias terms throughout our paper):
\begin{equation}
\label{eq:decision}
\frac{p_j(x)}{p_k(x)}=\frac{\exp (\mathbf{{w}}_j^\top {\boldsymbol{\phi}(\emph{x})} )}{\exp (\mathbf{{w}}_k^\top {\boldsymbol{\phi}(\emph{x})} )}=\exp [(\mathbf{{w}}_j - \mathbf{{w}}_k)^\top
\boldsymbol{\phi}(\emph{x})]
\end{equation}

\begin{figure}
\centering
\vspace*{-0in}
\subfigure[$\|\mathbf{w}_k\|_2 =\|\mathbf{w}_j\|_2 $]{\includegraphics[width=0.34\linewidth]{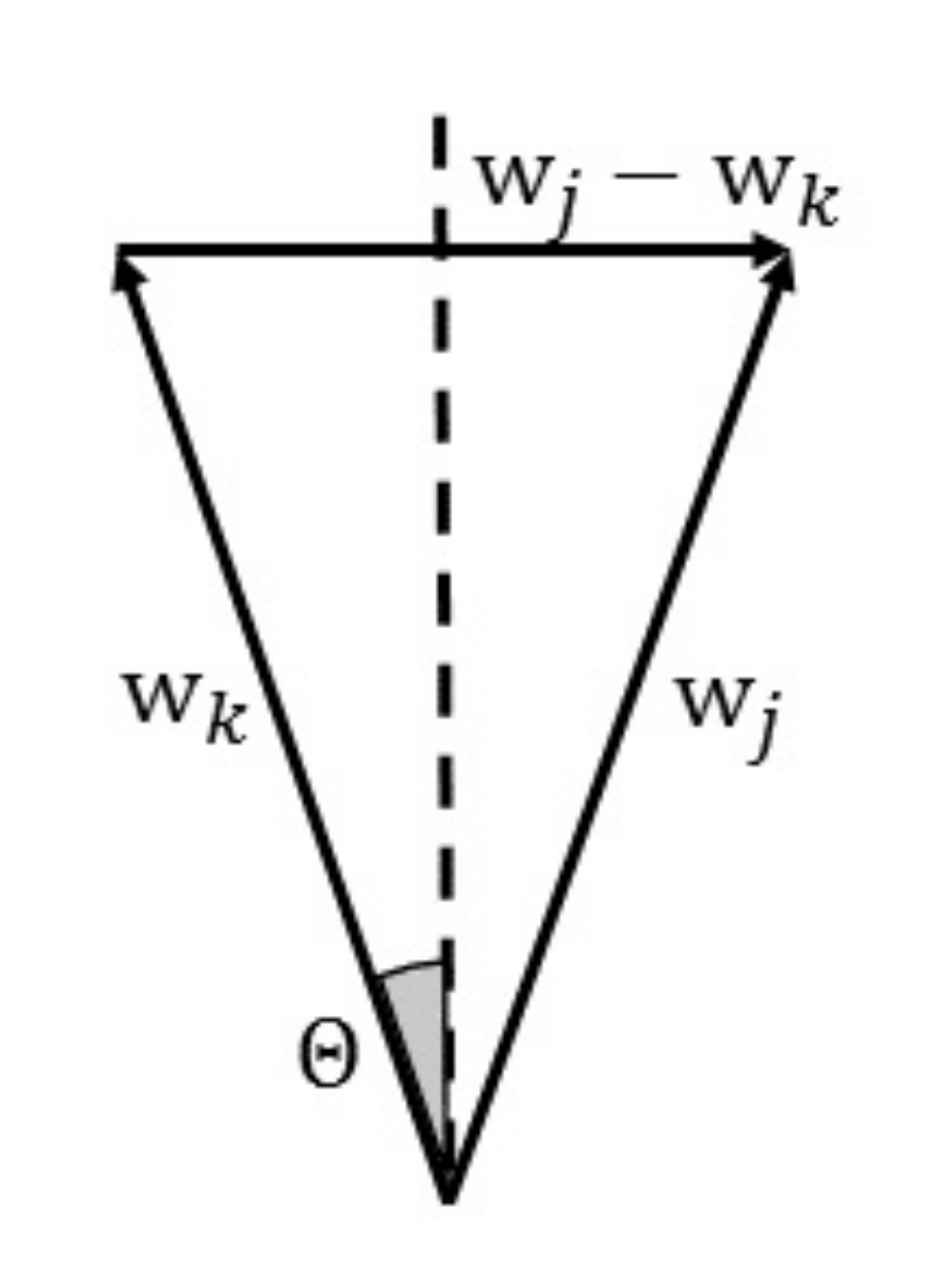}} \hspace{0cm}
\subfigure[$\|\mathbf{w}_k\|_2 <\|\mathbf{w}_j\|_2 $]{\includegraphics[width=0.36\linewidth]{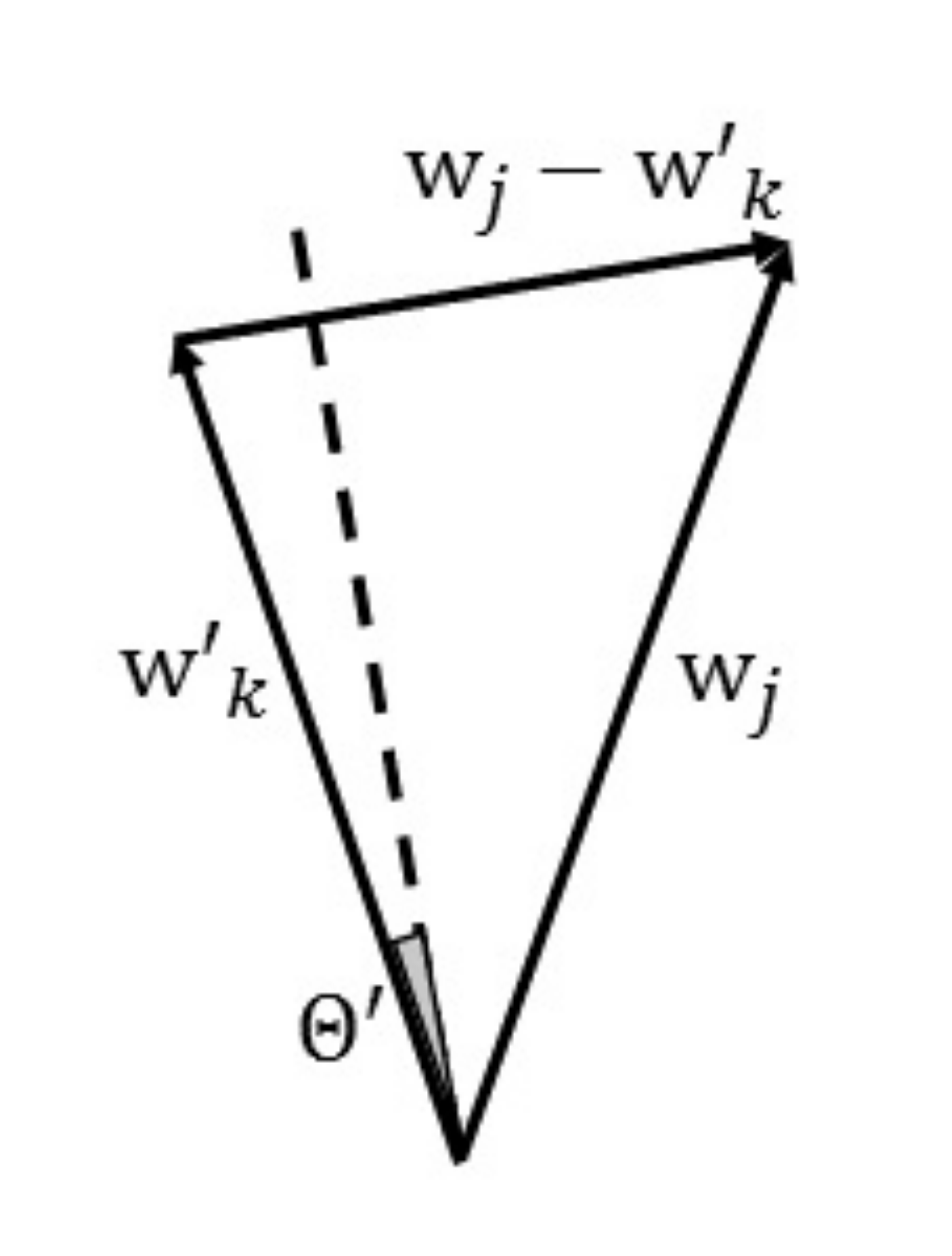}}
\caption{Relationship between the norm of $\mathbf{w}_k$ and the volume size of the partition for the $k$-{th} class. The dash line represents the hyper-plane (perpendicular to $\mathbf{w}_j-\mathbf{w}_k$) which separates the two adjacent classes. As shown, when the norm of $\mathbf{w}_k$ decreases, the $k$-{th} class tends to possess a smaller volume size in the feature space.}\vspace*{-2mm}
\label{fig:weight}
\end{figure}

As shown in Figure \ref{fig:weight}, the hyperplane to separate two adjacent classes $k$ and $j$ is perpendicular to the vector $\mathbf{w}_j - \mathbf{w}_k$. When the norm of $\mathbf{w}_k$ gets decreased, this hyperplane is pushed towards the $k$-{th} class, and the volume for the $k$-{th} class also gets decreased. As this property holds for any two classes, we can clearly see the connection of the norm of a weight vector and the volume size of its corresponding partition space in the feature space.

\vspace{1mm}\noindent{\bf Discussion}: There have been a lot of effort in integrating extra terms with cross-entropy loss to improve the feature generalization ability. The most similar version is the center loss in \cite{wen2016discriminative}, also known as 
the dense loss in \cite{Latha:dense} published during the same time. In center loss, the extra term is defined as 
\begin{equation}
\label{eq:center}
\mathcal{L}_c = - \sum_k\sum_{i \in C_k} \|\mathbf{c}_k - \boldsymbol{\phi}(\emph{x}_i) \|_2^2\, ,
\end{equation}
where $\mathbf{c}_k$ is defined as the \textit{class} center (might be dynamically updated as the approximation of the true class center due to implementation cost).

Our method is different from center loss from two perspectives. First, minimizing the cost function (Eq. \eqref{eq:center}) may lead to two consequences. While it helps reduce the discrepancy between $\boldsymbol{\phi}(\emph{x}_i)$ and its associated center $\mathbf{c}_k$, it also reduces the norms of $\boldsymbol{\phi}(\emph{x}_i)$ and $\mathbf{c}_k$. The second consequence is usually not good as it may hurt the classification performance. We did observe in our experiment that over training with center loss would lead to features with too small norms and worse performance compared with not using center loss (also reported in \cite{wen2016discriminative}). On the contrary, our loss term only considers the angular between $\boldsymbol{\phi}(\emph{x}_i)$ and $\mathbf{w}'_k$, and will not affect the norm of the feature. In our experiment section, we demonstrate that our method is not sensitive to the parameter tuning. 

Second, please note that we use the weight vector in Softmax $\mathbf{w}_k$ to represent the \textbf{classification} center, while in \eqref{eq:center}, the variable $\mathbf{c}_k$ is the \textbf{class} center. The major difference is that $\mathbf{w}_k$ is updated 
(naturally happens during minimizing $\partial \mathcal{L}_c$) using not only the information from the $k$-{th} class, but also the information from the other classes. In contrast, $\mathbf{c}_k$ is updated only using the information from the $k$-{th} class (calculated separately). More specifically, according to the derivative of the cross-entropy loss in Eq. \eqref{eq:crossentropy}, 
\begin{equation}
\label{eq:gcd}
\dfrac{\partial \mathcal{L}_s}{\partial \mathbf{w}_k}=\sum_i (p_{k}(\emph{x}_i)-t_{k,i}){\boldsymbol{\phi}(\emph{x}_i)}  \,, 
\end{equation}
the direction of $\mathbf{w}_k$ is close to the direction of the face features in the $k$-{th} class, and being pushed far away from the directions of the face features \textit{not} in the $k$-{th} class.

\subsection{Generative One-Shot Learning}

Conventional generative adversarial network (GAN) models attempt to synthesize fake data in the image space \cite{goodfellow2014generative}. Many kinds of improved variants incorporate class labels or latent information to obtain class conditional samples \cite{mirza2014conditional,oord2016conditional,reed2016generative}. To seek consistent distribution between real and synthesized data is a common problem to all the GAN-based approaches. In our proposed generative one-shot learning, we target at augmenting one-shot classes in the feature space. A general believe is that it is relatively easier to make the two distributions more consistent in the feature space rather than in the image space.

Given random noise $\emph{z}\in \mathbb{R}^{d_z}$, real feature $\boldsymbol{\phi}(\emph{x}) \in \mathbb{R}^{d_x}$ with its one-hot label $\emph{y} \in \mathbb{R}^{d_y}$. In the generator, the prior input noise $p_z(\emph{z})$, and one-hot label $\emph{y}$ are combined in joint hidden representation, and the adversarial training framework allows for considerable flexibility in how this hidden representation is composed. In the discriminator, $\boldsymbol{\phi}(\emph{x})$ and $\emph{y}$ are presented as inputs and to a discriminative function, where $\boldsymbol{\phi}(\cdot)$ denotes the feature extractor for sample $\emph{x}$. The objective function of a two-player minimax game is as follows:
 \begin{equation}
 \begin{array}{c}
 \mathcal{L}^f_d = \mathbb{E}[\mathrm{log}(1-D(G(\emph{z}|\emph{y})))]\\
 \\
 \mathcal{L}^r_d = \mathbb{E}[\mathrm{log}(D(\boldsymbol{\phi}(\emph{x})))]
 \end{array}
 \end{equation}
where the generator aims to make the generated features similar to real features, attempting to minimize $\mathcal{L}^f_d$; the discriminator aims to differentiate the real and fake features by maximizing $\mathcal{L}^r_d+\mathcal{L}^f_d$.

\begin{figure}[t]
\begin{center}
      \includegraphics[width=0.49\textwidth]{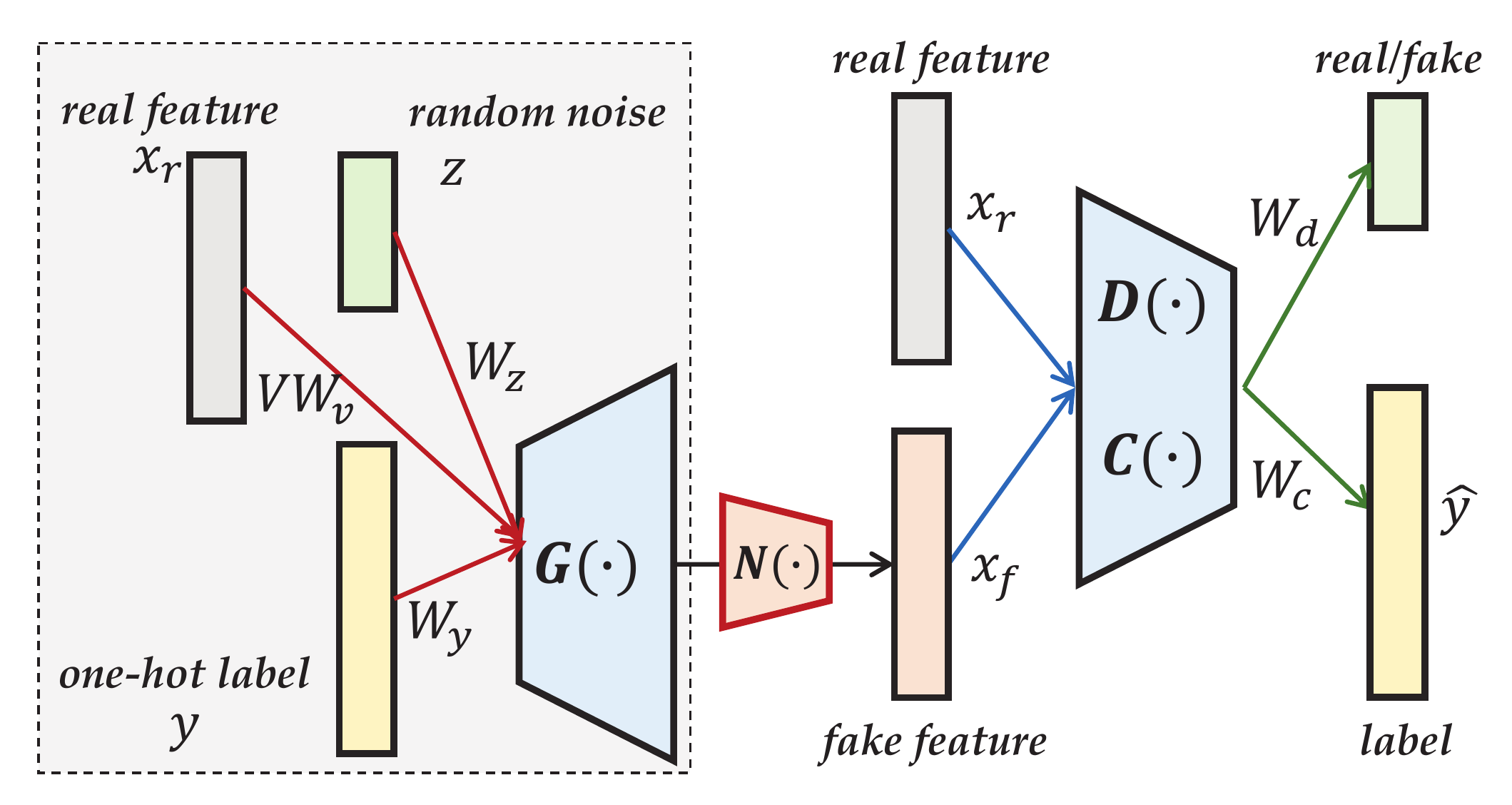}
  \vspace{-8mm}\caption{Illustration of generative one-shot face recognizer, where $z$ is the random noise vector, $\emph{y}$ is the one-hot label, $\emph{x}_r = \boldsymbol{\phi}(\emph{x})$ is the real feature, while $\emph{x}_f$ is the generated fake feature. $G(\cdot)$ is the generator with the input of random noise $\emph{z}$, original real feature $\emph{x}_r$, and one-hot label $\emph{y}$. The output of generator with normalization $N(\cdot)$ will achieve the fake feature $\emph{x}_f$. $D(\cdot)$ is the discriminator which aims to differentiate the real and fake features, while $C(\cdot)$ is the a general multi-class classifier.}\vspace{-6mm}\label{model}
\end{center}
\end{figure}

\subsubsection{Knowledge Transferable Generator}

As we know, the above conditional generative model does not involve the variance of base classes, which may not effectively adapt the knowledge from base classes to one-shot classes. We assume that base and one-shot classes share the same true distribution. While instances in the base class well-sample the true distribution, the one-shot instances are under sampling. By transferring the intra-class variance of a base class to a one-shot class, the feature distribution within the one-shot class can be enriched to be similar to base classes. The task then is to model the intra-class variance. By assuming the variance to be a multi-variate Gaussian, Cao et al. proposed a joint Bayesian model that takes the facial feature as the identity feature plus the intra-class variance \cite{cao2013practical}. From the observation of researchers, the identity feature is well-approximated by the feature center. Thus, the facial feature is represented as: $\boldsymbol{\phi}(\emph{x}_i) = \mathbf{c}_i + v_i$, where $c_i$ is $i$-th class feature center, $\boldsymbol{\phi}(\emph{x}_i)$ is one arbitrary sample in the class while $v_i$ represents the variance from arbitrary feature to the class center.

Different from Cao's assumption for the variance \cite{cao2013practical}, we completely rely on the data variance from the base classes. That follows the human cognitive process that we could adapt the previous knowledge (data variance from base classes) to learn new things (one-shot classes). Here we are not to model the variance across all the classes but rather to model the one-shot class variance. Given the observed feature distribution in base classes, we seek a parameterization to transfer the variance to the one-shot classes. Theoretically, any complete space decomposition method is valid here. In this paper we incorporate such knowledge transfer into the generator $G(\cdot)$ to simulate for effective features for one-shot classes. 

Specifically, we design the generator $G(\cdot)$ with the input of random noise $\emph{z} \in \mathbb{R}^{d_z}$, intra-classes variance of base classes $V \in \mathbb{R}^{d\times{d}}$ (later we will discuss how to obtain this matrix), and one-hot label $\emph{y} \in \mathbb{R}^{c}$. 

\renewcommand{\arraystretch}{1.5}
\begin{equation}\label{g1}
\begin{array}{rl}
  G(\emph{z}|V,\emph{y}) & = f_1(W_g\begin{bmatrix} \emph{z}\\ \emph{y} \end{bmatrix} + Vv_i) \\
   & = f_1([W_z, W_y]\begin{bmatrix} \emph{z} \\ \emph{y} \end{bmatrix} + Vv_i)\\
   & = f_1(W_z\emph{z}+W_y\emph{y} + Vv_i),
  \end{array}
\end{equation}
where $W_g = [W_z, W_y]$ while $W_z \in \mathbb{R}^{d\times{d_z}}$ and $W_y \in \mathbb{R}^{d\times{c}}$. $v_i \in \mathbb{R}^{d}$ is a coefficient vector to find the most related intra-class variances from the base class ($V$). To go deeper, we consider the variance of ons-shot classes could be represented by its nearby base classes. And $f_1(\cdot)$ is the element-wise activation function, e.g., ReLU function or Sigmoid function.

\vspace{1mm}\noindent{\bf Remark}: For the generator $G(\cdot)$, we attempt to synthesize meaningful data to augment the one-shot classes. The goal is to span the feature space of one-shot classes around its center features. Generally, we can calculate the mean feature of base classes as their centers, while the one-shot features for novel classes are usually not the centers and may be far away from their centers. From Eq. \eqref{g1}, we notice there are three parts used to generate the fake features. There two parts $W_z\emph{z}+W_y\emph{y}$ are from the conventional conditional generator \cite{mirza2014conditional}, which aims to seek two projections, one is for random noise and the other for conditional one-hot label. The third part $Vv_i$ follows a dictionary-based reconstruction format, that is, we hope to select the most relevant data variances from the base classes to synthesize one-shot feature. That is, we consider there are $d$ bases for class variance which can be well captured from base classes. Thus, the class variance for one-shot classes can be represented the combination of the bases in specific coefficients. On one hand, this term can adapt the knowledge of class variance from base classes to one-shot classes to synthesize meaningful data for one-shot classes. On the other hand, this term is able to better estimate the one-shot class centers since we only have one training sample in advanced so that we cannot obtain good one-shot class centers in the beginning. For simplicity, we adopt Principal Component Analysis (PCA) to parametrize the intra-class variance of regular classes, i.e., we first build a matrix $\mathcal{V} = [\boldsymbol{\phi}(\emph{x}_j^i)-c_i]_{j,i} \in \mathbb{R}^{d\times{n_b}}$ ($d \ll n_b$), then we select $d$ eigen-vectors with large eigen-values to represent the dictionary $V$.

However, our reconstruction term $Vv_i$ makes the sample dependent when optimizing $v_i$. Thus, we explore a projective dictionary to approximate $v_i$ with $W_v\boldsymbol{\phi}(\emph{x}_i)$. In this way, we only need to optimize a shared projective dictionary $W_v \in \mathbb{R}^{d\times{d}}$ for all classes. Hence, the generator can be reformulated as:
\renewcommand{\arraystretch}{1.5}
\begin{equation}\label{g2}
\begin{array}{c}
  G(\emph{z}|V,\boldsymbol{\phi}(\emph{x}),\emph{y}) = f_1\Big(W_z\emph{z}+W_y\emph{y} + VW_v\boldsymbol{\phi}(\emph{x})\Big),
  \end{array}
\end{equation}

Moreover, we hope $W_y\emph{y}$ could keep the class center information, while $W_z\emph{z}$ to compensate the residual information. Therefore, we initialize $W_y$ with the class-center features, and then we would get its class center by multiplying $W_y$ and its one-hot label $\emph{y}$. Specifically, the base part is initialized with the mean features of $c_b$ classes, while novel part is initialized with the available one-shot features. {\it Note that the one-shot class centers are not the real centers, so that we hope the intra-class variance part could facilitate to optimize the $W_y$, aseptically the novel part.} For random noise part, we initialize $W_z$ and $\emph{z}$ randomly. In this way, we can capture the data variation within base classes and adapt to generate more meaningful data for novel classes. Another thing is that the scale of $W_y\emph{y}$ is the same as that of $\emph{x}$, and thus, the scale would be improved if we add a random part $W_z\emph{z}$. Therefore, we add a normalization process to make the fake feature with the same scale to the real feature. Specifically, we define our normalization process as 
\begin{equation}
N(G(\emph{z}|V,\boldsymbol{\phi}(\emph{x}),\emph{y})) = \dfrac{\alpha G(\emph{z}|V,\boldsymbol{\phi}(\emph{x}),\emph{y})}{\|G(\emph{z}|V,\boldsymbol{\phi}(\emph{x}),\emph{y})\|_2},
\end{equation}
where $\alpha$ is the mean norm of real feature across all samples. 

\subsubsection{Joint Discriminator \& Classifier}

The generator $G(\cdot)$ attempts to augment more training data for one-shot classes, which is guided by the discriminator $D(\cdot)$. The goal of discriminator manages to differentiate the fake features and real features. In this way, two players could compete with each other to synthesize more effective data. Since we build the generator in the feature domain, we design one-layer fully-connected network to build the discriminator as follows: 
\begin{equation}\label{d1}
D(\emph{x}) = f_2(W_d\bar{\emph{x}}),
\end{equation}
where $W_d \in \mathbb{R}^{1\times{d}}$ and $f_2(\cdot)$ projects $\bar{\emph{x}}$ to a scale between 0 and 1. $\bar{\emph{x}}$ can be the real feature or generated feature. For real feature, the output of $D(\bar{\emph{x}})$ tends to be 1, otherwise 0.

Simultaneously, we target at building a general classifier $C(\cdot)$ across $c$ classes for both base classes and one-shot classes based on the real feature and synthesized feature. Specifically, we adopt the standard Softmax classifier with loss function $\mathcal{L}_s$ defined as (Eq. \eqref{eq:crossentropy}).

To sum up, we propose our generative one-shot learning model by incorporating generator, discriminator and classifier together into a unified framework. Specifically, there are two players, i.e., $D(\cdot)+C(\cdot)$ and $G(\cdot)$. For $D(\cdot)+C(\cdot)$, we train to maximize $-\mathcal{L}_s+\mathcal{L}^r_d-\mathcal{L}^f_d$, while $G(\cdot)$ is trained to maximize $-\mathcal{L}_s+\mathcal{L}^f_d$.

\vspace{2mm}\noindent{\textbf{Remark}}: For $C(\cdot)$, $W_c$ are the classifier parameters for both the base and novel classes. We initialize the classifier parameters trained on the base and novel dataset with the ResNet-34 deep features \cite{Resnet} (See detail in experiments). As known to all, a deep model training on base classes with many samples per class can achieve very promising results for base classes \cite{guo2017one}. That is, the classifier parameters are good enough for base classes recognition. The goal of one-shot learning is to improve the classifier parameters for one-shot classes. Hence, we hope the base classifier parameters to be similar to the pre-training one. We develop a square loss regularizer to constrain the base classifier not far away from its initialized one. In this way, not only can we update the classifier parameters for novel classes to enhance the classification ability, but also relax the classifier space for base classes, triggering the expansion of novel classifier space.

\vspace{2mm}\noindent{\textbf{Implementation}}: Training with one-shot classes usually results in a biased classifier. Our algorithm aims to correct this classifier bias by transferring variances from base classes to one-shot classes. For the generative one-shot learning model, we adopt the deep features from (ResNet-34) with loss (Eq. \eqref{eq:costfunction}) as the input and set the learning rate as $10^{-4}$ with the optimizer as Adam optimizer. We adopt leaky-relu and sigmoid activation functions for $G(\cdot)$ and $D(\cdot)$, respectively. Since GANs can be solved as a \textit{minimax} optimization problem, we first constrain the generator to optimize the discriminator, then fix the discriminator to update the generator. Thus, we iteratively update two neural networks until the model converges. This violation will help to correct the classifier bias and reshape the decision boundary (Figure \ref{gc}).

\section{Experimental Results}

In this section, we first introduce the one-shot face data as well as its feature representation process. Then we provide the one-shot evaluations with other comparisons to verify the effectiveness of our proposed model. Finally, we go deep and show some phenomena of our model.

We first train a general face representation model with the training images in the base set, and then train a multi-class classification model with the training images in both the base and novel sets. We list the experimental results in details in the following subsections. 

\subsection{One-Shot Face Dataset}

The face dataset\footnote{http://www.msceleb.org/challenge2/2017} used here is sampled from MS-Celeb-1M dataset \cite{guo2016ms}. In total, this dataset contains 21K people with 1.2M images, which is considerably larger than other publicly available datasets except for the MS-Celeb-1M dataset. To evaluate the one-shot challenge, we divide the dataset into base set ($20$K) parts, i.e., base set (20K people) and novel set (1K people). Since we want to build a general 21K-class classifier for both the base and novel classes, we hope our optimized classifier achieve promising performance on both sets, otherwise it is meaningless in the real-world applications.

In the base set, there are 20K persons, each of which having 50-100 images for training and 5 for test. In the novel set, there are 1000 persons, each with one image for training and 10 for test. The experimental results in this paper were obtained with 100K test images for the base set and 20K test images for the novel set. We focus on the recognition performance in the novel set, while monitoring the recognition performance in the base set to ensure that the performance improvement in the novel set does not harm the performance in the base set.

To recognize the test images for the persons in the novel set is a challenging task. The one training image per person was randomly preselected, and the selected image set includes images of low resolution, profile faces, and faces with occlusions. The training images in the novel set show a large range of variations in gender, race, ethnicity, age, camera quality (or evening drawings), lighting, focus, pose, expressions, and many other parameters.

\subsection{Face Representation Learning}
\label{sec:experiments-feature}
Learning good feature is the foundation of one-shot face recognition task. In order to evaluate the discrimination and generalization capability of our face representation model, we leverage the LFW \cite{LFWTech,LFWTechUpdate} verification task, which is to verify whether a given face pair (in total 6000) belongs to the same person or not. 

We train our face representation model (Eq. \eqref{eq:costfunction}) using the images in our base set (already published to facilitate the research in the area, excluding people in LFW by design) with ResNet-34 \cite{Resnet} with input faces' resolution as 224$\times$224. Specifically, we seek a 20K-class classifier using all the training images of the 20K persons in the base set. There are about 50-100 images per person in the base set. The wrong labels in the base set are very limited (less than 1\% based on manual check). We crop and align face areas to generate the training data\footnote{http://www.msceleb.org/download/lowshot}. Our face representation model is learned from predicting the 20K classes. We have tried different network structures and adopted the standard residual network with 34 layers \cite{Resnet} due to its good trade-off between prediction accuracy and model complexity. Features extracted from the last pooling layer are adopted as the face representation (512 dimensions).

The verification accuracy with different models are listed in Table \ref{table:lfwResults-o}. As shown, for the loss function, we investigated the standard cross entropy, cross entropy plus our WFB-loss term in Eq. \eqref{eq:costfunction}, the center loss in \cite{wen2016discriminative}, and the sphere face loss in \cite{sphereface}. For the WFB-loss, we set $\lambda$ in Eq. \eqref{eq:costfunction} as $0.1$. For the center loss, we tried different sets of parameters and found the best performance could be achieved when the balancing coefficient was $0.005$, as reported in the table. For the sphere face \cite{sphereface}, we noticed this paper very recently and only tried limited sets of parameters  (there are four parameters to be adjusted together). The parameters reported in the paper can not make the network converge on our dataset. The only parameter set we found to make the network converge leads to worse results, compared with the standard cross-entropy loss. Due to time constraint, for the other methods, we only report the results for some of them referring the numbers stated in the published corresponding papers. Please note that these methods use different datasets and different networks structures. 

\begin{figure}[t]
\begin{center}
     \includegraphics[width=0.49\textwidth]{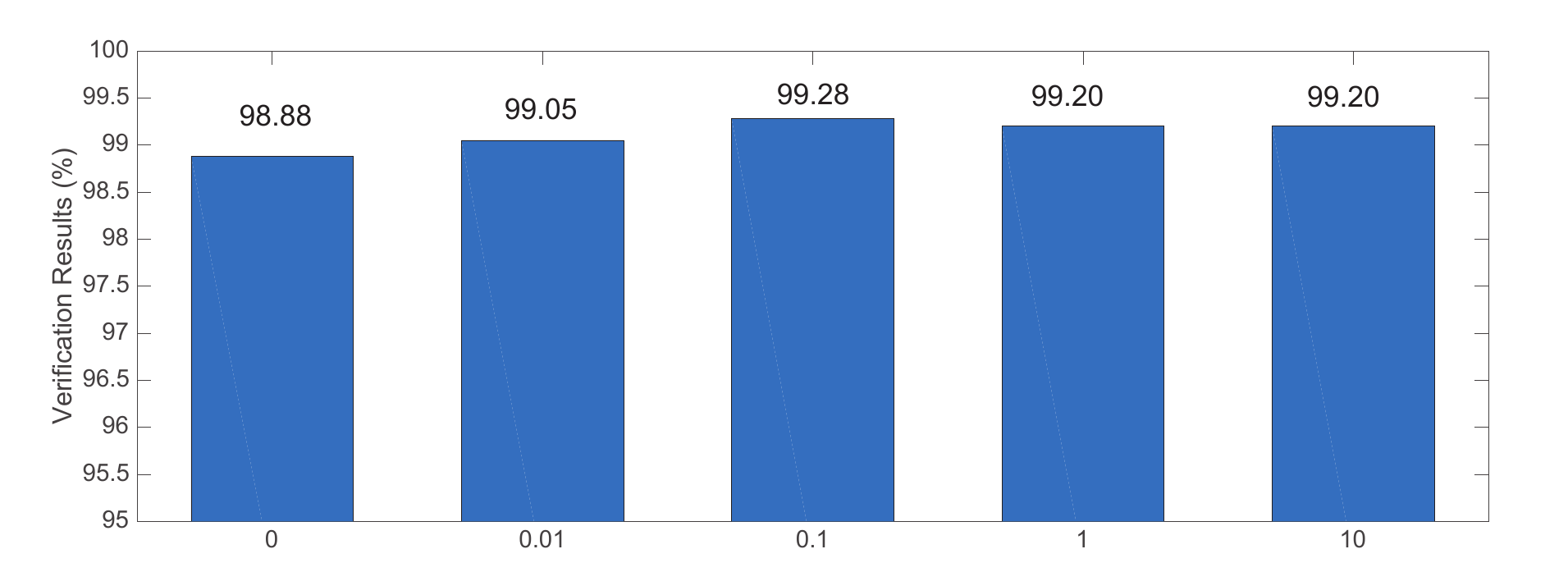}
\vspace{-5mm}\caption{LFW verification results obtained with different $\lambda$ for our WFB in Eq. \eqref{eq:cca}, where $\emph{x}$-axis denotes the values of $\lambda$.}\label{table:lambda}\vspace{-2mm}
\end{center}
\end{figure}

As shown in Table \ref{table:lfwResults-o}, we obtain the face representation model with the cutting-edge performance with the help of our WFB-loss term in Eq. \eqref{eq:costfunction}. We regard our model good enough to let us start to investigate the one-shot learning phase. We also tried different values of $\lambda$ in Eq. \eqref{eq:costfunction} and found our method is not sensitive to the choose of $\lambda$, shown in the Figure \ref{table:lambda}. Larger $\lambda$ means stronger regularizer applied. Note $\lambda = 0$ corresponds to no WFB-loss applied. From the results, we found $\lambda=0.1$ could generate better performance.

\begin{table}
\begin{center} \caption{
LFW verification results obtained with models trained with our published base set. All the models use ResNet-34 \cite{Resnet} as the feature extractor. For the sphere face, please refer to our paper for explanation (fail to converge).}\label{table:lfwResults-o}
\begin{tabular}{lcc}
   \Xhline{1pt}
Methods &Network&Accuracy \\
\hline
Cross entropy only  & 1 & $98.88\%$  \\
Center face \cite{wen2016discriminative}  & 1 & $99.06\%$  \\
Sphere face \cite{sphereface} & 1 & $-.--\%$  \\
Cross entropy + WFB in Eq. \eqref{eq:costfunction} (ours)  & 1 & $\mathbf{99.28}\%$  \\
   \Xhline{1pt}
\end{tabular}
\end{center}

\end{table}

\begin{table}
\begin{center} \caption{
For reference, LFW verification results (partially) reported in peer-reviewed publications. Different datasets and network structures were used.}\label{table:lfwResults-r}
\begin{tabular}{lccc}
    \Xhline{1pt}
Methods &Dataset&Network&Accuracy \\
\hline
JB \cite{cao2013practical}& Public & -- & $96.33\%$  \\

Human & -- & -- & $97.53\%$ \\

DeepFace\cite{vgg_face}  & Public & 1 & $97.27\%$ \\

DeepID2,3 \cite{Xiaoou_Deep2,Xiaoou_Deep3}& Public & 200 & $99.53\%$  \\

FaceNet \cite{schroff2015facenet} & Private & 1 & $99.63\%$ \\

Center face \cite{wen2016discriminative}  & Private & 1 & $99.28\%$  \\

Center face \cite{wen2016discriminative}  & Public & 1 & $99.05\%$  \\

Sphere face \cite{sphereface} & Public & 1 & $99.42\%$  \\
Our WFB in Eq. \eqref{eq:costfunction}  & Public & 1 & $\textbf{99.73\%}$  \\
  \Xhline{1pt}
\end{tabular}
\end{center}
\end{table}

\subsection{One-shot Face Recognition}

In phase two, we train a $21,000$-class classifier to recognize the persons in both the base set and the one-shot set. Since there is only one image per person for training in the one-shot set, we repeat each sample in the one-shot set for $100$ times through all the experiments in this section. In order to test the performance, we apply this classifier with $120,000$ test images consists of images from the base or one-shot set. We focus on the recognition performance in the novel set while monitoring the recognition performance in the base set to ensure that the performance improvement in the novel set does not harm the performance in the base set. 

To recognize the test images for the persons in the novel set is a challenging task. The one training image per person was randomly preselected, and the selected image set includes images of low resolution, profile faces, and faces with occlusions. We provide more examples in the supplementary materials due to space constraint. The training images in the novel set show a large range of variations in gender, race, ethnicity, age, camera quality (or evening drawings), lighting, focus, pose, expressions, and many other parameters. 
Moreover, we applied de-duplication algorithms to ensure that the training image is visually different from the test images, and the test images can cover many different looks for a given person.

\begin{table}
\begin{center}\caption{Coverage at Precisions = 99\% and 99.9\% on the one-shot set, where our generative model significantly improves the coverage at precision 99\% and 99.9\%.}\label{c@p}
\renewcommand{\arraystretch}{1.4}
\begin{tabular}{ccc}
  \Xhline{1pt}
  Method & C@P=99\% & C@P=99.9\%\\
  \hline
    Fixed-Feature & 25.65\% & 0.89\%\\
    SGM \cite{hariharan2016low} & 27.23\% & 4.24\%\\
    Update Feature & 26.09\% & 0.97\% \\
    Direct Train & 15.25\% & 0.84\%\\
    Shrink Norm \cite{guo2017one} & 32.58\%  & 2.11\%\\
    Equal Norm \cite{guo2017one} & 32.56\% & 5.18\%\\
    Up Term \cite{guo2017one} & 77.48\% & 47.53\% \\
    \hline
  Hybrid \cite{Wu_2017_ICCV} & $92.64\%$ & N/A \\
  Doppelganger \cite{ICCV-W-Doppelganger} & $73.86\%$ & N/A  \\
  Generation-based \cite{ICCV-W-Generation} & $61.21\%$ & N/A \\
    \hline
    Ours & 94.98\% & 83.94\% \\
  \Xhline{1pt}

\end{tabular}
\end{center}\vspace{-2mm}
\end{table}

\begin{figure}[t]
\begin{center}
     \includegraphics[width=0.49\textwidth]{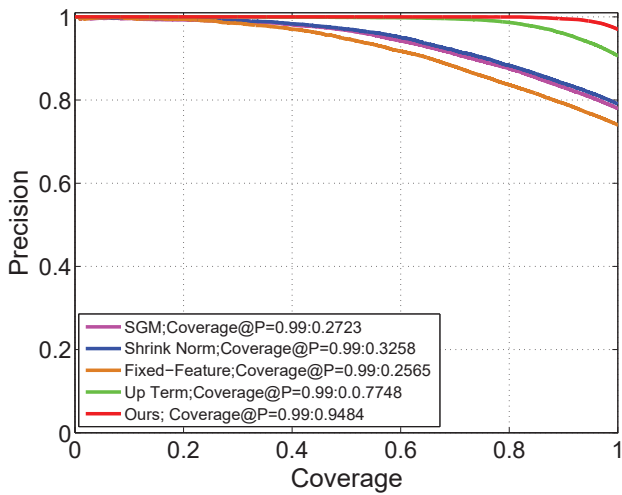}
\vspace{-3mm}\caption{Precision-Coverage curves of five methods on the one-shot set, where our model achieves a very appealing coverage@precision=99\%.}\label{c@p_curve}\vspace{-5mm}
\end{center}
\end{figure}

We compare with the following algorithms:
\begin{itemize}
  \item \textbf{Fixed-Feature}: updates the feature extractor and only train the Softmax classifier with the feature extractor provided by phase one.
  \item \textbf{Updated Feature}: fine-tunes the feature extractor simultaneously when we train the Softmax classifier in phase two. The feature updating does not change the recognizer's performance too much.
\item \textbf{SGM} \cite{hariharan2016low}: is known as squared gradient magnitude loss, is obtained by updating the feature extractor during phase one using the feature shrinking method.
\item \textbf{Shrink norm} \cite{guo2017one}: adopts $L_2$-norm to shrink classifier parameters, which is one typical strategy to handle insufficient data problem efficiency.
\item \textbf{Equal norm} \cite{guo2017one}: is a weight regularizer, which constrains the classifier parameters of both novel and base classes to the same value.
\item \textbf{UP Term} \cite{guo2017one}: is a weight regularizer, which only enforces the classifier parameters of the novel classes to the same value.
\end{itemize}

All the methods are based on a 21K-class classifier (trained with different methods). Note that we boost all the samples in the novel set for 100 times for all the methods, since the largest number of samples per person in the base set is about 100.

\begin{figure}[!t]
\begin{center}
     \includegraphics[width=0.5\textwidth]{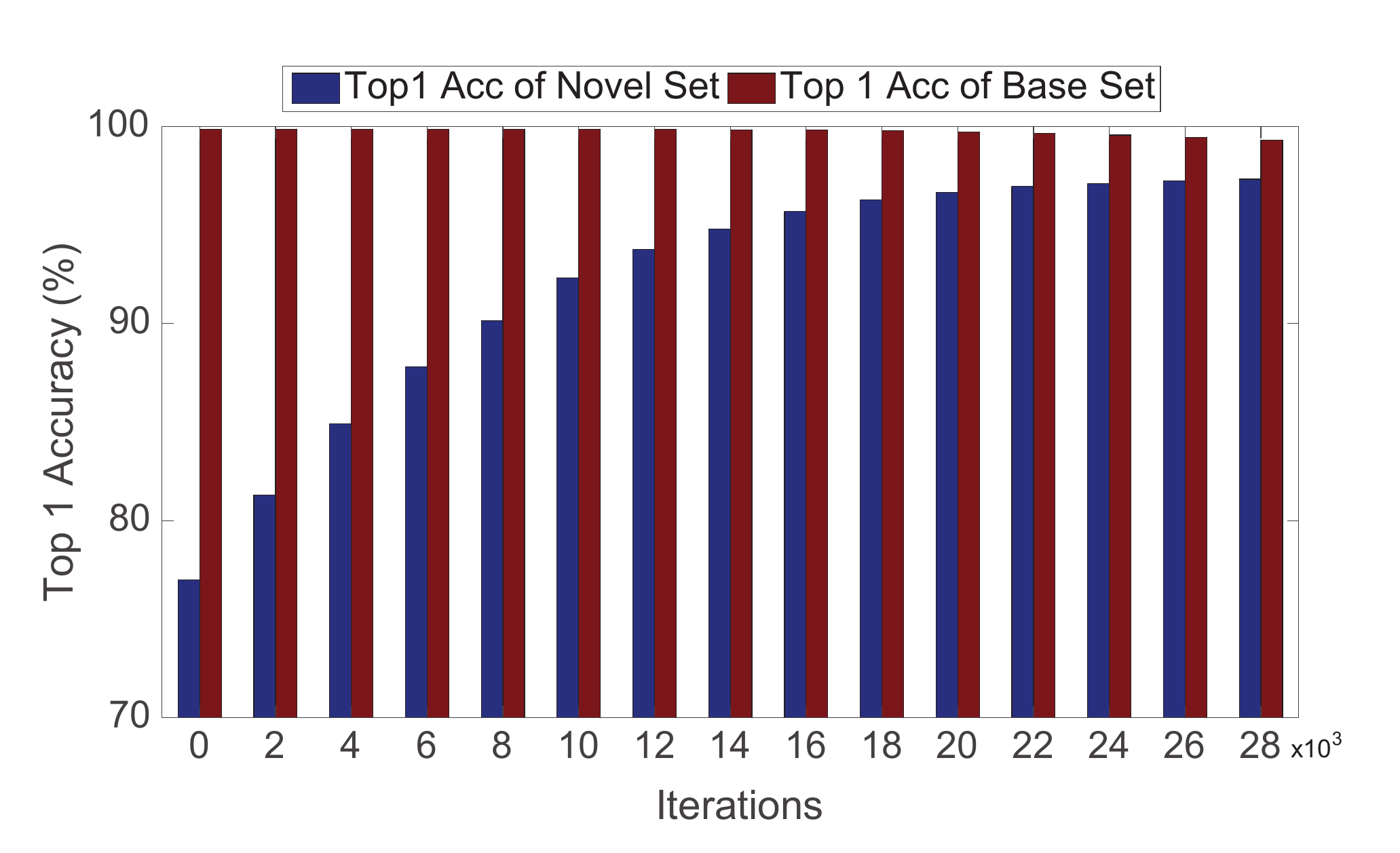}
\vspace{-3mm}\caption{Top1 accuracy (\%) of base set and novel set with different iterations, where we notice that our model could significantly improve the Top1 accuracy for the novel classes while keeping a very promising Top1 accuracy for the base classes.}\label{acc_curve}\vspace{-2mm}
\end{center}
\end{figure}

\begin{figure*}[!t]
\begin{center}
     \includegraphics[width=0.95\textwidth]{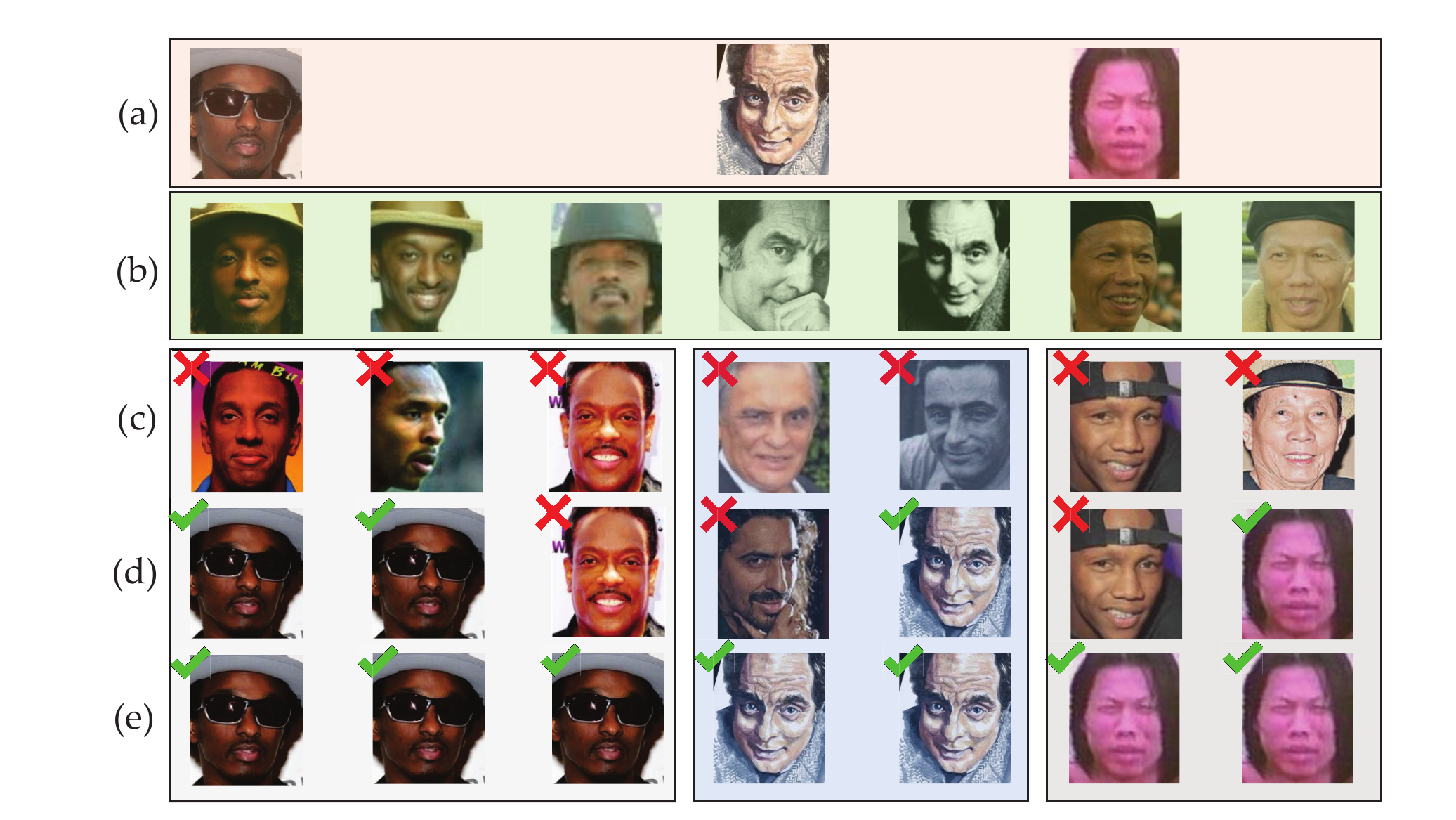}
\vspace{-2mm}\caption{Face retrieval results, where row (a) denotes the three challenges one-shot training faces, i.e., occlusion, sketch, low-resolution. Row (b) represents the test images, while the the bottom three rows show the recognized results of three models, i.e., (c) $k$-NN, (d) Softmax, and (e) Our generative model. }\label{visualization}\vspace{-6mm}
\end{center}
\end{figure*}

\begin{figure*}[!t]
\begin{center}
     \includegraphics[width=0.98\textwidth]{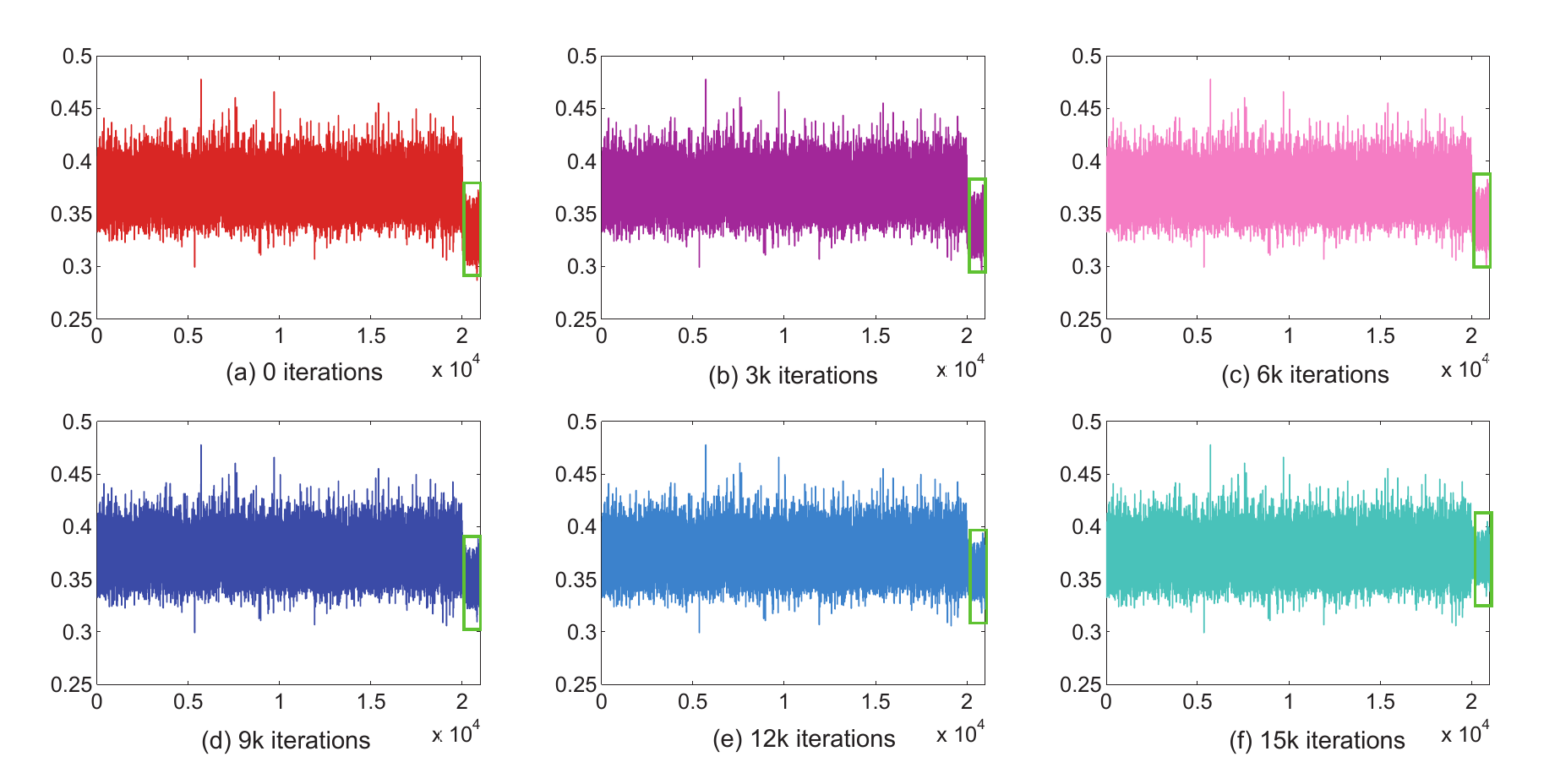}
\vspace{-5mm}\caption{Norm of the classifier weight vector $w$ for each class in $W_c$. The $\emph{x}$-axis is the class index. The rightmost 1000 classes on the x-axis correspond to the persons in the novel set. As shown in the figure, with more iterations from (a) to (f), $\|w\|_2$ for the novel set tends to have similar values as that of the base set (Green bounding box denotes the weights for one-shot classes). This promotion introduces significant performance improvement. }\label{softmax_weight}\vspace{-5mm}
\end{center}
\end{figure*}

The experimental results of our method and the alternative methods are listed in Table \ref{c@p}. We adopt coverage rate at precision $99\%$ and $99.9\%$ as our evaluation metrics since this is the major requirement for a real recognizer \cite{guo2017one}. As shown in the table, our method significantly improves the recall at precision $99\%$ and $99.9\%$ and achieves the \textbf{best} performance among all the methods. Unless numbers reported by other papers (Hybrid, Doppelganger, and Generation-based), the face feature extractor was trained with cross entropy loss. 

Compared with the Fixed-Feature, SGM method obtains around {2\%} improvements in recall when precision is $99\%$, while {4\%} improvements when precision requirement is $99.9\%$. The gain for face recognition by feature shrinking in \cite{hariharan2016low} is not as significant as that for general image. The reason might be that the face feature is already a good representation for faces and the representation learning is not a major bottleneck. Note that we did not apply the feature hallucinating method as proposed in \cite{hariharan2016low} for fair comparison and to highlight the contribution of model learning, rather than data augmentation. To couple the feature hallucinating method (may need to be modified for face) is a good direction for the next step.

Our model significantly improves the one-shot classification, preserving base classification at a very promising performance. Specifically, as shown in Table \ref{c@p}, our generative model improves the coverage@precision=$99\%$ and coverage@precision=$99.9\%$ significantly. Moreover, we notice that our model can achieve the state-of-the-art performance without any external data by comparing the competitors in the low-shot challenge \footnote{http://www.msceleb.org/leaderboard/c2}. This verifies our generative model is able to synthesize very effective features to alleviate the one-shot classification.

The coverage at precision $99\%$ on the base set obtained by using any classifier-based methods in Table \ref{c@p} is $100\%$. The Top1 accuracy on the base set obtained by any of these classifier-based methods is $99.80\pm{0.02}$\%. Thus, we do not report them separately in the table. That verifies that our generative model could synthesize meaningful one-shot samples to boost classifier space for one-shot classes.

\subsection{Face Retrieval Results}

We select three typical one-shot training cases (Figure \ref{visualization}), i.e., low-solution, sketch, occlusion, to quantitatively show the performance of different models. We compare with $k$-nearest neighbor classifier ($k$-NN) ($k=1$), Softmax, and our one-shot generative model. All the models are input with the pre-trained deep ResNet-34 features with our newly designed loss (Eq. \eqref{eq:costfunction}).

From the results (Figure \ref{visualization}), we observe that our model can well handle these three challenging cases and recognize these persons correctly. $k$-NN cannot correctly recognize the testing images of these three persons, which results from that the testing images are quite different from the one-shot training image. Softmax can retrieve some correct ones, which shows more promising results than $k$-NN. Hence, we consider $k$-NN is not suitable in one-shot face classification in large-scale dataset. Our model can significantly handle those three challenging cases, which results from the generation of effective data in facilitating the classifier learning. 

\subsection{Property Analysis}

First of all, we evaluate the Top1 accuracy of the base set and novel set with the model optimization. From the results \ref{acc_curve}, we observe that the Top1 accuracy of one-shot set is significantly improved from $77.01\%$ to $96.82\%$. This shows that our model enhances the classification for one-shot classes by spanning the feature space. We further notice that the classification accuracy for the base set is hurt somehow, but very slightly. That demonstrates our generative model can learn a good general classifier, which is much practical in real-world scenarios.

Secondly, we present more information for the classifier to deeply understand why our model can improve the one-shot classification. Specifically, we have a $c$-class classifier, with each class weight vector $w$. Thus, we evaluate the norm of each class weight vector to see the variations of these information. From the results (Figure \ref{softmax_weight}), we notice that from (a) to (f) with more iterations' optimization, the norms of the novel classes are triggered to similar distribution as the base classes (f). Actually, (a) shows the results of the initialized parameters obtained from Softmax trained on ResNet-34 deep features. That is the reason we consider why our model significantly improves the one-shot classification, since we boost the mean of novel classes to be similar to base classes. Such phenomenon is also obtained in \cite{guo2017one}, where the assume the norm of classifier weight is related to the classifier space. 

\section{Conclusions}

In this paper, we proposed a generative framework for one-shot face recognition, where we attempted to synthesize more effective augmented data for one-shot classes by borrowing the data variation of base set. Specifically, generative learning was jointly incorporated in the general classifier training for both the base and novel classes. Thus, more effective fake data were generated for the one-shot classes to enrich the data space of one-shot classes. Furthermore, a discriminator was designed to guide the face data generation to mimic the data variation of base classes and adapt to generate novel classes. Experiments on a large-scale one-shot face benchmark showed that our model could significantly improve the performance of one-shot classification, while keeping the promising classification ability for the base set. Studying how to generate data in image space will lead to another topic – a trivial face image synthesis work cannot lead to a convincing conclusion.

\bibliographystyle{IEEEtran}
\bibliography{paper_ref}

\end{document}